\documentclass{article}

\usepackage[T1]{fontenc}
\usepackage[utf8]{inputenc}

\usepackage{geometry}
\usepackage{tipa}
\usepackage{lingmacros}

\usepackage{synttree}
\branchheight{4ex}

\usepackage{natbib}

\usepackage{amsfonts,amssymb,amsmath}

\title{Morphonette: a morphological network of French}

\author{Nabil Hathout\\
  Université de Toulouse\\
  Nabil.Hathout@univ-tlse2.fr}

\date{}

\begin{document}
\maketitle
\begin{abstract}
  This paper describes in details the first version of Morphonette, a new French morphological resource and a new radically lexeme-based method of morphological analysis.  This research is grounded in a paradigmatic conception of derivational morphology where the morphological structure is a structure of the entire lexicon and not one of the individual words it contains.  The  discovery of this structure relies on a measure of morphological similarity between words, on formal analogy and on the properties of two morphological paradigms: morphological derivational families and morphological derivational series.
\end{abstract}

\section{Paradigmatic derivational morphology}
\label{sec:parad-deriv-morph}

\begin{figure*}[t]
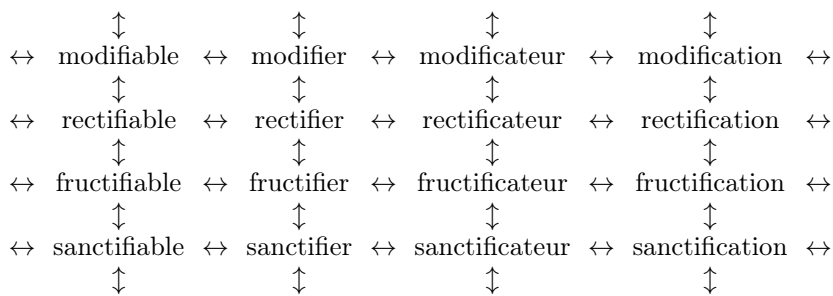

  \centering\begin{tabular}{c@{~~}c@{~~}c@{~~}c@{~~}c@{~~}c@{~~}c@{~~}c@{~~}c}
& $\updownarrow$ && $\updownarrow$ && $\updownarrow$ && $\updownarrow$ &\\
$\leftrightarrow$ & modifiable  &$\leftrightarrow$ & modifier & $\leftrightarrow$ & modificateur & $\leftrightarrow$ & modification & $\leftrightarrow$\\
& $\updownarrow$ && $\updownarrow$ && $\updownarrow$ && $\updownarrow$ &\\
$\leftrightarrow$ & rectifiable & $\leftrightarrow$ & rectifier & $\leftrightarrow$ & rectificateur & $\leftrightarrow$ & rectification & $\leftrightarrow$\\
& $\updownarrow$ && $\updownarrow$ && $\updownarrow$ && $\updownarrow$ &\\
$\leftrightarrow$ & fructifiable  & $\leftrightarrow$ & fructifier & $\leftrightarrow$ & fructificateur & $\leftrightarrow$ & fructification & $\leftrightarrow$\\
& $\updownarrow$ && $\updownarrow$ && $\updownarrow$ && $\updownarrow$ &\\
$\leftrightarrow$& sanctifiable & $\leftrightarrow$ & sanctifier & $\leftrightarrow$ & sanctificateur & $\leftrightarrow$ & sanctification & $\leftrightarrow$\\
& $\updownarrow$ && $\updownarrow$ && $\updownarrow$ && $\updownarrow$ &
\end{tabular}

  \caption{The morphological network of the French lexicon is made up of derivational families and derivational series. Families and series are connected by morphological analogies.}
  \label{fig:maillage-morphologique}
\end{figure*}

The starting points of this research are the fundamental ideas of lexeme-based morphology \citep{aronoff94.morphology-itself}: only lexemes are signs (i.e.\ atomic units); affixes are merely phonologial marks; the construction of the meaning and of the form of a derived word are distinct processes.  It is grounded in a conception of derivational morphology where words do not have a morphological structure and where this structure is a level of organization of the lexicon.  This organization is based on the semantic, formal and categorical relations that hold between the words memorized in the lexicon  \citep{bybee95.regular-morphology}.  Among these relations, analogies play a prominent role because they allow the emergence of the morphological paradigms. An analogy is a quaternary relations $a:b::c:d$ that holds between the members of a quadruplet $(a,b,c,d)$ such that $a$ is to $b$ as $c$ is to $d$.   Morphological derivational analogies holds between the members of two types of paradigms : morphological derivational families and morphological derivational series.  This can be illustrated with an analogy such as \emph{duplication}~: \emph{duplicateur}~:: \emph{unification}~: \emph{unificateur}\footnote{%
`duplication', `duplicator', `unification', `unifier'%
} where we can see that \emph{duplication} and \emph{duplicateur} belong to the same derivational family and that it goes the same for \emph{unification} and \emph{unificateur}.  This conception enables us to redefine the morphological analysis task, which aims to make explicit the morphological paradigms of the lexicon instead of decompose the individual words into morphemes.  This organization is illustrated in figure~\ref{fig:maillage-morphologique}.  The analysis of a given word then consists in identifying its position in the morphological structure of the lexicon.  For instance, the word 
 \emph{rectificateur} `recitifier' is not analyzed as in (\ref{ex:rectificateur-tree}) but as a member of the derivational family which contains \emph{rectifiable}, \emph{rectifier} `rectify', \emph{rectifieur} `recitifier', \emph{rectification}, \emph{rectificatif} `corrective', etc. and of the derivational series which contains \emph{certificateur} `certifier', \emph{fructificateur} `which bears fruits', \emph{modificateur} `modifier', \emph{sanctificateur} `sanctifier', etc.   These two sets can be seen as the morphological coordinates of \emph{rectificateur}.
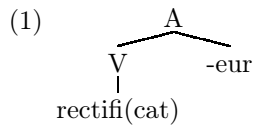
\enumsentence{\label{ex:rectificateur-tree}
\evnup[3pt]{\synttree[A [V [rectifi(cat)]] [-eur]]}
}

The objective of the present research is twofold: first, we propose a radically lexeme-based method of morphological analysis capable of providing the morphological derivational structure of the lexicon; second, we have computed this structure for a significant fragment of a large-coverage lexicon of French.  This resource, Morphonette, will soon be made available to the public.

A morphological network solves several problems posed by the morphematic approach such as the treatment of words such as \emph{concevoir} `conceive', \emph{décevoir} `deceive', \emph{percevoir} `perceive', \emph{recevoir} `receive' or \emph{consister} `consist', \emph{désister} `desist', \emph{persister} `persist', \emph{résister} `resist' where it is difficult to determine the status to the \mbox{\emph{con-},} \emph{dé-}, \emph{per-}, \emph{re-}, \emph{-cevoir} or \emph{-sister} sequences.  The paradigmatic approach is also capable of bringing words such as \emph{furieux} `furious' and \emph{curieux} `curious' into the same lexical derivational series despite the fact that \emph{furieux} has a derivational base, \emph{furie} `fury', while the current lexicon of French contains no word that could serve as a base to \emph{curieux}.  The dissociation of the construction of meaning and form allows us to easily treat allomorphy, suppletion and phenomena such as interfixation that one observes in \emph{goutte} `drop' $\rightarrow$ \emph{gouttelette} `droplet' or \emph{triste} `sad' $\rightarrow$ \emph{tristounet} `gloomy' described by \cite{plenat2005.decalage-ette}.

The network illustrated in figure~\ref{fig:maillage-morphologique} is actually made up of analogies.  For instance, \emph{fructificateur}:\emph{fructification} participates in analogies with \emph{modificateur}:\emph{modification}, \emph{rectificateur}:\emph{rectification}, \emph{sanctificateur}:\emph{sanctification}.  Similarly, \emph{fructificateur}:\emph{rectificateur} forms analogies with \emph{fructifier}:\emph{rectifier}, \emph{fructification}:\emph{rectification}, \emph{fructifiable}:\emph{rectifiable}.  Gathering all theses analogies poses a serious problem of complexity.  For instance, for a lexicon of $97~010$ entries such as the \emph{Trésor de la Langue Française} (TLF) word list, the number of quadruplets to be tested is on the magnitude of $10^{19}$.  This number is theoretically $10^{20}$ but it can be divided by $8$ by taking advantage of the permutations described in (\ref{eq:1}) where ${\cal L}$ is a set of representations of the lexical units. 
\addtocounter{equation}{1}
\begin{eqnarray}
  \label{eq:1}
  \lefteqn{\forall (a,b,c,d)\in {\cal L}^4, a:b::c:d \Rightarrow a:c::b:d \wedge b:a::d:c\ \wedge }\\
  &&  b:d::a:c \wedge c:a::d:b \wedge c:d::a:b \wedge d:a::c:b \wedge d:c::a:b \nonumber
\end{eqnarray}
For the construction of the Morphonette network, we have used the phonological representations of the TLF headwords instead of their written forms, so reducing the size of the lexicon to $83~082$ entries and the number of quadruplets to be checked to $6\cdot10^{18}$.

The solution we adopted for the complexity problem consists in using the measure of morphological similarity proposed by \cite{hathout2008.textgraphs3}.  This measure enables us to select for a given entry $w$ the words that are most likely to form analogies with $w$, namely the members of the derivational family and series of $w$ (see section~\ref{sec:morph-simil}).  The second problem we have had to solve is the actual verification of the analogies.  We have used the same algorithm as \cite{hathout2008.textgraphs3}.  Inspired by the one of \cite{lepage98.COLING}, this algorithm allows us to check whether a formal analogy holds between four words without having to cut them into morphemes.  Notice that this algorithm may exceptionally fail to find some analogies.  Another algorithm, proposed by \cite{stroppa2005.these}, does not suffer from this drawback.  However, we did not use it because its complexity is in $o(n^4)$ while the former has a complexity in $o(n^2)$ and because these exceptional failures are largely compensated by the number and the redundancy of the collected analogies.  The construction of Morphonette poses a third problem, namely the exclusion of the formal analogies that are not morphologically valid such as \emph{constituable}~: \emph{constant}~:: \emph{restituable}~: \emph{restant}\footnote{%
`constitutable', `constant', `restitutable', `remaining'%
}.  We relied on the structure of the morphological graph to eliminate them, namely on the fact that series contain large numbers of words, that they are clusters with highly connected members and that series are connected to each others by large numbers of edges which form analogies.  Notice that the series of the lexicon too form a cluster.

The remainder of the paper is organized as follows.  In Section~\ref{sec:morph-simil}, we present the measure of formal similarity and the morphological neighborhoods where the analogies are looked for. Section~\ref{sec:formal-analogy} outlines the verification of the formal analogies.  In Section~\ref{sec:morph-netw}, we describe in detail the bootstrapping algorithm we have used for the construction of this first version of Morphonette.  The resource is presented in Section~\ref{sec:morphonette-0.1}.  Section~\ref{sec:related-works} discusses some related works and finally, Section~\ref{sec:concl-direct-furth} offers a short conclusion.

\section{Morphological similarity}
\label{sec:morph-simil}

We have used the measure of morphological similarity proposed by \cite{hathout2008.textgraphs3} for the construction of Morphonette.  This measure brings closer the words that share large numbers of very specific formal and semantic features: the more features the words share and the more specific these features are, the closer they are.  The measure is calculated by means of a bipartite graph where the words are connected to their features.  The neighbors of a word $w$ are identified by spreading an activation initiated at the vertex that represents $w$. First, the activation is uniformly spread toward the features of $w$. Then, in the second step, the activation located on the features is uniformly spread toward the words that possess these properties.  The level of activation obtained by a word $x$ after the propagation is an estimation of the morphological relatedness between $w$ and $x$.  The spreading is simulated by means of a classical random walk algorithm, that is by multiplying the stochastic adjacency matrix of the bipartite graph.

The measure originally proposed by \cite{hathout2008.textgraphs3} uses both formal and semantic properties, the latter being $n$-grams of words extracted from the TLF definitions.  We did not retain them here because they are not informative enough.  Another difference with \cite{hathout2008.textgraphs3} is the use of phonetic transcriptions instead of word forms.  We have used the LIA\_PHON phonetizer of \cite{bechet2001.liaphon} in order to transcribe the word forms into sequences of phonemes in Mbrola format.  Each phoneme is encoded as two characters as shown in the examples in (\ref{ex:rep-mbrola}). 
\addtocounter{enums}{1}
\enumsentence{\label{ex:rep-mbrola}
  \evnup[3pt]{
    \begin{tabular}[t]{ll}
      constant & kkonssttan\\
      constituable & kkonssttiittuuaabbllee\\
      restant & rraissttan\\
      restituable & rraissttiittuuaabbllee
    \end{tabular}
  }
}
 The beginning and the end of the words are marked by \texttt{\#\#}. The morphological similarity is then estimated by associating with each word the set of all the sequences of $3$ phonemes or more.  For instance, the sequences which describe the word \emph{constant} are presented in (\ref{ex:rep-mbrola-contant}).
\enumsentence{\label{ex:rep-mbrola-contant}
{\small\texttt{\#\#kkon}~~ \texttt{kkonss}~~ \texttt{onsstt}~~ \texttt{ssttan}~~ \texttt{ttan\#\#}\\
\texttt{\#\#kkonss}~ \texttt{kkonsstt}~ \texttt{onssttan~} \texttt{ssttan\#\#}\\
\texttt{\#\#kkonsstt}~~ \texttt{kkonssttan}~~ \texttt{onssttan\#\#}\\
\texttt{\#\#kkonssttan\#\#}}
}
Figure~\ref{fig:voisins-fructifier} presents the nearest neighbors of \emph{fructifier} `bear fruit'.  If we omit \emph{sanctifier} `sanctify', \emph{rectifier} and \emph{présanctifier} `presanctify', we see that the members of the derivational family of \emph{fructifier} all appear at the beginning of the list and that the end gathers the members of its derivational series.
\begin{figure}[t]
  
\textbf{fructifier fructifiant fructificateur fructification fructifiant fructifère} \emph{sanctifier rectifier présanctifier} \textbf{fructivore} fructidorien fructidorienne fructidoriser fructidor \textbf{fructueusement fructueux fructuosité fructose} obstructif constructif instructif désobstructif destructif instructif autodestructif \textbf{usufructuaire infructueusement} sanctifiant sanctifiable rectifieuse rectifieur rectifiant rectifiable \emph{transsubstantifier substantifier stratifier cimentifier certifier savantifier refortifier ratifier présentifier pontifier plastifier notifier nettifier mortifier mythifier mystifier quantifier}

  \caption{The $50$ nearest neighbors of \emph{fructifier} `bear fruit'.  The members of the derivational family are in bold face and the ones of the derivational series are in italic.}
  \label{fig:voisins-fructifier}
\end{figure}

\section{Formal analogy}
\label{sec:formal-analogy}

The measure of morphological similarity enables us to determine a morphological neighborhood for each word $w$. This neighborhood gathers a large part of the members of the derivational family and series of $w$.  These members are precisely the ones with which $w$ can form morphological analogies. In this way, we can reduce drastically the search space for analogies, as proposed in \cite{hathout2008.textgraphs3}.  For instance, if we limit the search to the $100$ first neighbors of each word, the number of quadruplets to be checked for a lexicon of $83~082$ entries drops to $10^{10}$.  This number can be further reduced by using two heuristics based on the properties (\ref{eq:2}) and (\ref{eq:3}).
\addtocounter{equation}{2}
\begin{equation}
  \label{eq:2}
  \forall (a,b,c,d)\in {\cal L}^4, a:b::c:d \Rightarrow l(a)-l(b)=l(c)-l(d)
\end{equation}
where $l(x)$ is the number of phonemes in $x$.
\begin{eqnarray}
  \label{eq:3}
  \lefteqn{\forall (a,b,c,d)\in {\cal L}^4, a:b::c:d \Rightarrow  }\\
&& (c(a)=c(b)\wedge c(c)=c(d))\vee(c(a)=c(c)\wedge c(b)=c(d))\nonumber
\end{eqnarray}
where $c(x)$ is the morphosyntactic tag of $x$.  Morphonette uses the Grace tag set \citep{rajman97.grace-morpho-syntaxique}.  These heuristics divide the total number of quadruplets to be checked by $50$.  $2\cdot10^8$ quadruplets have therefore been checked and $4.2\cdot10^6$ formal analogies have been collected.  In order to further improve the quality of these analogies, we have only kept the ones where a formal analogy also holds for the written forms.  This additional condition eliminates phonetic analogies such as \emph{paissant}~: \emph{abaissant}~:: \emph{paye}~: \emph{abeille}\footnote{%
`grazing', `lowering', `pay, `bee'%
}.  The number of analogies actually used for the construction of the first version of Morphonette is $3.9\cdot10^6$.  The set of these analogies is closed under the permutations described in (\ref{eq:1}). Let ${\cal A}$ be this set.

The analogies in ${\cal A}$ have been found by using the same technique as the one of \cite{hathout2008.textgraphs3} which consists in computing an analogical signature for each of the pairs of words $(a,b)$ and $(c,d)$ of a quadruplet $(a,b,c,d)$.  The analogical signature of a pair of words $(a,b)$ describes a path in their edit lattice, that is a sequence of string edit operations.  $(a,b,c,d)$ is an analogy if the two signatures are identical. This method fails to detect some analogies such as (\ref{ex:do-doable}).\footnote{%
  We thank Philippe Langlais who pointed out this problem to us.%
}
\addtocounter{enums}{2}
\enumsentence{\label{ex:do-doable}
\emph{do}~: \emph{doable}~:: \emph{read}~: \emph{readable}
}
These failures being exceptional and the analogies highly redundant, it is always possible to recover the relations $a:b$ and $c:d$ and then the entire analogy $a:b::c:d$.  Notice that the algorithm of  \cite{stroppa2005.these} is able to identify (\ref{ex:do-doable}), but it has a complexity in $o(n^4)$.  It is obviously not adapted to our needs given the number of quadruplets we have to check.

\section{Morphological network}
\label{sec:morph-netw}

Morphonette has been constructed by using a bootstrapping algorithm.  We first selected an initial seed, ${\cal M}_0$, composed of the most reliable morphological relations and then complemented it iteratively with relations induced by ${\cal M}_0$.  More specifically, the $3.9\cdot10^6$ collected analogies were used to define a weighted graph ${\cal G}=(V,E,w)$ where $V$ is a set of vertices, namely the set of the headwords of the TLF, $E=\{(a,b)\in V\times V/ \exists a:b::c:d \in {\cal A}\}$ a set of edges and $w:E\rightarrow \mathbb{N}$ a weight function such that $\forall e\in E,w(e)=|\{a:b::c:d \in {\cal A}/(a,b)=e\}|$.  ${\cal G}$ being build from formal analogies, the words represented by the vertices are mainly connected to members of their derivational families on one hand and to members of their derivational series on the other.  The main objective of the construction of Morphonette is to set apart these two types of relations and to select a set of relations with almost no error.  This is because ${\cal A}$ contains formal analogies such as \emph{destructeur}~: \emph{structural}~:: \emph{descripteur}~: \emph{scriptural}\footnote{%
 `destructor', `structural', `descriptor', `scriptural'%
} which induce morphologically invalid edges, namely \emph{destructeur}:\emph{structural} and \emph{descripteur}:\emph{scriptural}.

The relations between members of the same family and members of the same series can be partially set apart on the basis of the categorical features of the words: two words that belong to the same series have identical morphosyntactic tags.  As a result:
\addtocounter{equation}{1}
\begin{equation}
  \label{eq:4}
  \forall a:b::c:d \in {\cal A}, c(a)\neq c(b) \Rightarrow \phi(a,b) \wedge \phi(c,d) \wedge \sigma(a,c) \wedge \sigma(b,d)
\end{equation}
\begin{equation}
  \label{eq:5}
  \forall a:b::c:d \in {\cal A}, c(a)\neq c(c) \Rightarrow \phi(a,c) \wedge \phi(b,d) \wedge\sigma(a,b) \wedge \sigma(c,d)
\end{equation}
where $\phi(x,y)$ is true iff $x$ and $y$ belong to the same derivational family and $\sigma(x,y)$ is true iff $x$ and $y$ belong to the same derivational series.  However, this criterion does not allow us to type the edges of analogies where $c(a)=c(b)=c(c)=c(d)$ such as \emph{développeur}~: \emph{développement}~:: \emph{enveloppeur}~: \emph{enveloppement}\footnote{%
 `developer', `development', `enveloper', `envelopment'%
} which holds between four masculine singular nouns.  The statements (\ref{eq:4}) and (\ref{eq:5}) can be used to define a type function $\tau$ of the analogies in ${\cal A}$:
\begin{equation}
  \label{eq:6}
  \tau(a:b::c:d)= \left\{%
    \begin{array}{ll}
      \mbox{f} & \mbox{if $c(a)\neq c(b)$}\\
      \mbox{s} & \mbox{if $c(a)\neq c(c)$}\\
      \mbox{u} & \mbox{othewise}
    \end{array}%
  \right.
\end{equation}
We can then define the subset of $E$ made up of the edges which connect words which may be in the same family:
\begin{equation}
  \label{eq:7}
  {\cal F}=\{(a,b)\in E / \exists a:b::c:d\in {\cal A}, \tau(a:b::c:d)\in\{\mbox{f},\mbox{u}\}\}
\end{equation}

The partial typing of the edges in ${\cal G}$ can be refined on the basis of two structural characteristics of the morphological network.  These characteristics allows us to select a subgraph of ${\cal G}$ with  the most reliable morphological relations only:
\addtocounter{enums}{4}
\enumsentence{\label{ppt-series-1}
  Derivational series are large sets.
}
\enumsentence{\label{ppt-series-2}
  Derivational series are clusters.
}

The characteristic (\ref{ppt-series-1}) allows us to identify reliable family relations. This is because two words $a$ and $b$ which belong to the same family normally participate to one analogy with each of the members of the series of $a$ and of $b$.  Series being large sets, the weight $w(e)$ of an edge $(a,b)$ connecting members of the same family is normally high.  In other words, the number of analogies which contain a given edge can be used identify the ones which reliably connect members of the same family.  For instance, a threshold of $10$ can be used to select a set which only contains family edges.  Let ${\cal F}_0=\{e\in {\cal F}/ w(e) \ge 10\}$ be this set.  We can then rely on ${\cal F}_0$ to identify relations between words which belong to the same series:
\addtocounter{equation}{2}
\begin{equation}
  \label{eq:8}
  \forall a:b::c:d \in {\cal A}, (a,b)\in{\cal F}_0\Rightarrow \sigma(a,c) \wedge \sigma(b,d)
\end{equation}
${\cal F}_0$ can therefore be used  to extract a subgraph ${\cal G}_0$ from ${\cal G}$ composed with serial relations induced by the reliable familial relations in ${\cal F}_0$:
\begin{eqnarray}
  \label{eq:9}
  {\cal S}_0 & = &\{ (a,c) \in E / \exists (a,b) \in {\cal F}_0 \mbox{ and } \exists a:b::c:d \in {\cal A} \}\\
  {\cal G}_0 & =& {\cal F}_0\cup{\cal S}_0
\end{eqnarray}

The structure we get is actually more complex. This is because one word $c$ can belong to the series of a word $a$ when $a$ is in a relation with a member $b$ of its family but not belong to the series of $a$ when $a$ is in a relation with another member $b'$.  For instance, \emph{artificiel} `artificial' belongs to the same series as \emph{officiel} `official' and \emph{troisième} `third' when it is in a relation with \emph{artificiellement} `artificially' but it is only in the same series as \emph{officiel} when in a relation with \emph{artificialiser} `artificialize'.  In the first case, \emph{artificiel}:\emph{artificiellement} forms analogies with \emph{officiel}:\emph{officiellement} `officially' and \emph{troisième}:\emph{troisièmement} `thirdly'; in the second, \emph{artificiel}:\emph{artificialiser} only forms an analogy with \emph{officiel}:\emph{officialiser} `officialize' but none with a pair having \emph{troisième} as its first member.  In other words, each entry belong to as many distinct sub-series as there are members in its family.  Thus, the morphological structure of the lexicon consists in a set of \textbf{filaments} of the form  $(a,b,\mathit{series}(a,b))$ where $a$ is an entry, $b$ a member of its family and $\mathit{series}(a,b)=\{c\in V/\exists a:b::c:d \in {\cal A}\}$ the sub-series of $a$ when we consider its relation with $b$.  Actually, the filaments of an entry $a$ are just a representation of the set of the analogies which contain $a$.\footnote{
Let us notice that filaments could be defined in a dual manner from the derivational series.  In this case, a filament of an entry $a$ is a triplet $(a,b,\mathit{family}(a,b))$ where $b$ is a member of the series of $a$ and $\mathit{family}(a,b)$ is the sub-family of $a$ when we consider its relation with $b$.  Both types of filaments being equivalents, we have used the first one because it yields a more compact description of the graph.%
}  Filaments are illustrated in figure~\ref{fig:filament}.

The characteristic (\ref{ppt-series-2}) is then used to enhance the selection of the most reliable edges in ${\cal G}$ starting from the most central serial relations.  Even if almost all the familial relations in ${\cal F}_0$ are correct, we need to eliminate the ones that may yield errors when the initial seed is extended, and especially the ones that connect distinct families.  These connections primarily concern compounds such as \emph{zoophilie} `zoophilia' which belong to the family of \emph{zoologie} `zoology' \emph{zoophobie} `zoophobia', etc.\ and to the one of \emph{anthropophilie} `anthropophilia', \emph{bibliophilie} `bibliophilia', etc.\ depending on whether we consider its radical is \emph{zoo} or \emph{philie}.  In this case, we eliminate the relation between \emph{zoophilie} and \emph{anthropophilie} by relying on the fact that \emph{zoophilie} has predominantly words ending in \emph{-philie} in its series and that these words do not have words starting with \emph{zoo-} in their series.  Put differently, the words starting with \emph{zoo-} are not well connected within the central cluster of the series of \emph{zoophilie}.  We classically measure the clustering coefficient of a word $c$ within the series of a word $a$ by the ratio of the number of triangles to the number of triples which contain the edge $(a,c)$ \citep{watts98.nature}.  Let $s_0(a)=\{c \in V / (a,c) \in S_0\}$ be the series of $a$.  Then the number of triples formed by $a$ and one word $c\in s_0(a)$ is  $|s_0(a)|-1$.  The number of triangles that a word  $c \in s_0(a)$ form with other members of $s_0(a)$ is $|(s_0(a)\setminus \{c\}) \cap (s_0(c)\setminus \{a\})|$.  A threshold of $0.66$ has been used for the construction of Morphonette.  It allows us to reduce the series to their most central clusters.  For series $s_0(a)$, this cluster can be defined as in (\ref{eq:10}).
\begin{equation}
  \label{eq:10}
  s'_0(a)=\{c \in s_0(a)/ \frac{|(s_0(a)\setminus \{c\}) \cap (s_0(c)\setminus \{a\})|}{|s_0(a)|-1}\ge 0.66\}
\end{equation}
This reduction is then used to remove from ${\cal F}_0$ the edges $(a,b)$ such that $\mathit{series}(a,b)\cap s'_0(a)=\emptyset$. The resulting graph is the initial seed ${\cal M}_0$.

${\cal M}_0$ is then iteratively extended until a fixed-point is reached.  At step $i$, we generate all the formal analogies induced by the transitive closures of the families of ${\cal M}_i$.  These analogies $a:b::c:d$ consists of to pairs $(a,b)$ and $(c,d)$ such that $\exists (t_1,t_2)\in {\cal T}_i\times{\cal T}_i, (a,b)\in t_1\times t_1\mbox{ and } (c,d)\in t_2\times t_2$ where ${\cal T}_i$ is the transitive closure of the families of ${\cal M}_i$.  We then reduced the graph induced by these analogies to its intersection with ${\cal G}$ and added this extension to ${\cal M}_i$ in order to yield ${\cal M}_{i+1}$.  We actually impose to the extension an additional condition: for $i\ge2$, only the filaments with a sub-series of $5$ words or more are kept.  The fixed-point is reached in 8 iterations.  The Morphonette network is the constructed by merging ${\cal M}_8$ with ${\cal G}_0$.

\section{Morphonette~0.1}
\label{sec:morphonette-0.1}

This first version of Morphonette comprises $29~310$ entries and $96~107$ filaments, and therefore the same number of familial relations.  The number of distinct families has not been computed.  The network contains $1~160~098$ serial relations, that is $12$ per filament in average.  These numbers can be compared with the ones of ${\cal G}$, the graph from which this network has been extracted.  ${\cal G}$ comprises $75~832$ entries, $816~922$ filaments (that is $10$ per entry in average, against only $3$ in Morphonette) et $2~343~059$ serial relations (that is less than $3$ per filament).  Morphonette therefore already covers about 40\% of the entries of the lexicon.  Figure~\ref{fig:filament} presents an excerpt of this resource consisting of three filaments of the noun \emph{gazouillarde} `twittering female'.
\begin{figure}[t]
  
  \begin{description}
  \item[gazouillarde] \emph{gazouillage}\\
    cafouillarde grenouillarde  vasouillarde
  \item[gazouillarde] \emph{gazouillement}\\
    braillarde geignarde grognarde 
  \item[gazouillarde] \emph{gazouiller}\\
    citrouillarde douillarde grenouillarde rouillarde souillarde vadrouillarde vasouillarde
  \end{description} 

  \caption{Three filaments of the entry \emph{gazouillarde} `twittering female'}
  \label{fig:filament}
\end{figure}

A first estimation of the quality of Morphonette has been performed by manually checking $200$ filaments randomly extracted from the network.  Only one erroneous relation has been found between \emph{pension} and \emph{pensif} `pensive' which puts the precision above 99\%, if confirmed by a more thorough evaluation.  Even if \emph{pension} and \emph{pensif} are etymologically related, there is nowadays no semantic relation between them.  However, \emph{pension} and \emph{pensif} participate to a large number of formal analogies which wrongly put \emph{pension} in the extended series of deverbal nouns ending in \emph{-ion}.  The loss of the semantic relation between \emph{pension} and \emph{pensif} can only be detected on the basis of semantic information. But Morphonette~0.1 has been constructed only from the formal properties of the TLF headwords.

Morphonette~0.1 also contains some errors due to formal accidents such as the relation between  \emph{dégrimer} `remove the make-up' and \emph{dégression} `degression' which belongs, from a formal point of view, to the series of \emph{déprimer}:\emph{dépression}\footnote{%
 `depress', 'depression'%
}, \emph{comprimer}:\emph{compression}\footnote{%
 `compress', `compression'%
}, etc.  Once again, the use of semantic knowledge should be the best way to find out and eliminate this type of errors. Another line of investigation would be to generalize the notion of analogy to sets of three pairs or more in order to determine the invariants of the sub-series.

Another difficulty we will have to address is the treatment of homonyms and homographs. For instance, the four meanings of \emph{fraise} (`strawberry', `mesentery', `ruff', `drill') induce four distinct derivational families even if the three latter meanings are etymologically related.  In Morphonette~0.1 these families are confused.  We will use the homonyms numbers in the TLF entries and the semantic information present in the definitions to separate them in future versions of Morphonette.

\section{Related Works}
\label{sec:related-works}

From a theoretical point of view, this work belongs to a framework related to the Network Morphology of \cite{bybee95.regular-morphology}, to the Surface-to-Surface Morphology of \cite{burzio99.surface-to-surface-morphology}, and to emergentist approaches of \cite{aronoff94.morphology-itself}, \cite{albright2002.PhD} or \cite{goldsmith2006.NLE}.

The construction of Morphonette uses a bootstrapping algorithm in order to extend an initial reliable seed.  This technique has often also been used in computational morphology, for instance by \cite{goldsmith2006.NLE} or by \cite{bernhard2006.FinTAL}.  However, our method differs from these ones because it is fully lexeme-based and does not make use of morpheme nor contain any representation of them.  Morphological regularities emerge from a very large set of analogies.  Gathering of this set is one of contributions of the work presented in this paper. It was made possible through the use of the measure of morphological similarity of \cite{hathout2008.textgraphs3}.  This measure was inspired by work on small words by \cite{gaume2002.JETAI}.  Our method is also close to the ones of \cite{yarowsky2000.ACL} and \cite{baroni2002.sigphon6} where the words are not decomposed into morphemes. Both make use of string edit distance to identify formal similarity between words.  Our work is also close to the one by \cite{stroppa2005.CoNLL}, \cite{langlais2009.EACL} and \cite{lavallee2009.morphochallenge} who use formal analogies to analyze words morphologically and to translate them.

The Morphonette network could also be compared to the morphological families constructed by \cite{xu98.ACM}, \cite{gaussier99.ACL} or \cite{bernhard2009.morphochallenge} among others.  It is also very close to Polymots, a manually-constructed morphological lexicon \citep{gala2010.LREC}.  Polymots and Morphonette are complementary since the former primarily contains short words while the latter mainly contains long words because of the criteria we have used to select the morphological relations.

With respect to these related works, the main contribution of Morphonette is first the generation of a collection of more than $4$ millions formal analogies and the exploitation of the structural properties of the morphological graph in order to set apart the familial and the serial relations.

\section{Conclusion and directions for further research}
\label{sec:concl-direct-furth}

We have presented in this paper Morphonette, the first morphological network of French.  This network is constructed without decomposition of the words into morphemes.  The method we have used rely on the structural properties of a graph of morphological relations build from a collection of almost $4$ millions formal analogies.  Morphonette is made up of filaments which are composed of an entry, a member of its derivational family and derivational sub-series of similar words.  It allows us to redefine the morphological analysis task which does not aim to decompose words into morphemes but aims to identify their derivational families and series by means of a set of filaments.

Morphonette will soon be distributed under Creative Commons licence.  A thorough evaluation of its relations will also be carried out shortly.  A second version of this resource will be developed by designing a measure of semantic relatedness able to differentiate between homonyms, to spot out the formal accidents and to identify allomorphy and suppletion.  This measure will be based on the relations in Morphonette~0.1 which will be used to select the semantic properties and relations which are the most informative from a morphological point of view.

\bibliographystyle{apalike2.bst}
\bibliography{biblio}

\end{document}